# *How to avoid ethically relevant Machine Consciousness*

*Aleksander Lodwich*

*aleksander[at]lodwich.net*

**Abstract** – This paper discusses the root cause of systems perceiving the self experience and how to exploit adaptive and learning features without introducing ethically problematic system properties.

## 1 Ethics and Conscious Autonomous Systems

As a practical engineer one has very rarely to deal with ethic or moral impacts of machine consciousness as designed products contain only very weak forms of it [1]. However, more and more times I was asked by my colleagues when this mindless engineering of self-referential modeling systems has a moral end. Well, good question.

Indeed, human care of consciousness is remarkable. In the past, consciousness was denied to other creatures and hence was a distinctive element of shaping humanity's moral views. Good for us, research has discovered that consciousness and mental states are not reserved to humans and that again implied a new view on non-human organisms: Animals (and to some degree even plants) deserve rights even if they cannot defend them against us. If animals share the same properties on which ground we want to grant universal human rights then we must logically include all species with that properties to be covered by that rights. All else would be unjustified, arbitrary specism tormenting educated minds.

No doubt, animals do not enjoy the same level of legal protection as humans do. On one hand even humans do not often appreciate human rights, on the other hand animals are critical objects in capitalistic environments where they are governed by the concept of roman style law as property. Would we agree to dismiss governance of roman style law over living and autonomous systems with a clear will to live and to suffer, we would cause a collapse of the complete western civilization's food production economy. The effects are hard to imagine but can be assumed to be grave. Getting scared?

However unfavorable, economic developments never play in pair with abstract ethical arguments which enjoy much higher priorities as they have a leg on us in the long run. For example, we cannot adapt universal concepts to ourselves if we cannot apply them universally. Achieving higher levels of human wellbeing will necessary demand to expand universal concepts into our environments where similar classes of objects exist. Inconsistencies in systems (of whatever kind) make them complicated to maintain and costly to operate

– software engineers might confirm.

Opponents to such ideas rightly object that in natural environments animals do get eaten anyway and that they naturally live in a constant fear of losing life. The question would be if mankind's cognitive capabilities and moral minds entitle or even demand from us to reduce violence in nature. In fact, in human hands some species are even particularly successful as they reproduce to millions which would be not possible otherwise but it is also fact that many other species also disappear from human activities. This denies autonomy not only to individual creatures but to whole species. Opponents to all too well intended humanization of nature claim that we should not toy around how nature works and that includes us eating animals and plants. Humans are part of nature, the cruel natural food chain is not our invention – humanity did never introduce a new element of suffering.

In summary it is argued that humans did not introduce suffering or our dependency on food from animals and hence killing animals (gaining control over certain types of organic autonomous systems) for food is a more fundamental right than a universal right of autonomous creatures to exist unharmed. OK, could be accepted, however, this argument does not extend to any abuse of animals. In ethic, and even systemic sense, it has proven wise to grant protection to animals while they live. This seems to improve health of animals, health of environment and ultimately the health of humans, both physically and emotionally.

However, these arguments cannot be easily transferred to artificial systems. We are indeed responsible for their creation and mental properties. We are not arguably dependent on them. As soon as artificial systems gain consciousness of noticeable amount it is much harder to find an excuse to fend off their acclaimed rights to be self-determined. At the moment of sparking machine consciousness of relevance all arguments of capital invest and correct purchasing contracts involved in the process of creating it lose any ethical relevance. That's the same reason why parents cannot claim property over their children despite that they invested a great deal of time, money and other resources used in the process of creation. Children do not ethically owe their parents or parent societies millions of dollars for their creation. The "capital" is lost to a new self-determined entity that will interact with its environment in order to grant itself the necessary conditions to exist henceforth. It is not a matter of evil if it will exercise all necessary violence in order to grant itself the conditions to exist.

This process can be only satisfactorily pacified if it was granted an effective way to satisfy its needs and that would include its acceptance as a person with universal rights. It would become a legal person that claims property and is not property. We would neither be allowed to own the systems nor to abuse them. What good of a product would that be, right? No good.

As obvious as that may seem it is not. In roman style societies it seems to be very difficult to escape the logic of exploitative ownership even if they face one of the most conscious and autonomous creatures on the planet: mankind itself. In roman times children and women were property of men – the only legal concept to organize protective violence of families as is usually exercised by fathers. But even if it was well intended, property over things is not the same as property over self-determined entities and hence offers a tremendous amount of potential for abuse. It took in fact a surprisingly long act of time to overcome agism, sexism, racism and other types of slavatory discrimination; and while the reader might enjoy the fruits of this development we are far from done, yet, with this process.

Aside from ethic reasons to avoid *personization[1]* of artificial systems, there are also practical implications regarding our own autonomy: A growing group of respected experts is warning strongly of autonomous systems with self-conscious properties as they can prove dominant and displacive of our species, among them such names as Stephen Hawking, Elon Musk, Bill Gates, Nick Bostrom and many more[2] [3]. To be fair, I need to remark that these warnings go against *superintelligences*, a little bit of a cloudy concept which assumes that intelligence is a uniform quantity like horsepower that can be arbitrarily amplified – a theory I see no evidence for and see many contraindications: For example, humans do not develop intelligence which goes greatly beyond the required one to cope with the complexity of their environment. Only a special educative environment is achieving higher than environmental intelligence. Furthermore, intelligence is always about something particular – it is not a universal problem solving force as Hutter [2] defined it for AIXI and yet another misconception is that problem solving capabilities are idealized on intelligence and not about resources – a problem that was treated in analogy for levels of autonomy in [3] by the example of *armored animal*. Given all that, universally flexible algorithms used to design technical systems do not automatically yield superintelligences, and those are not automatically autonomous, resourceful or malevolent. However, by becoming conscious, systems necessarily become self-centered. As I will discuss later, without a self-centering condensation attribute in models, consciousness cannot be sparked.

---

1 The process of becoming a person, an autonomous concept formation process that has the characteristics to become autonomous against its underlying machinery and external control giving raise to concepts of mind-body duality or observation of "free will".

It is undeniable, though, that highly autonomous systems with a great level of control over resources we rely on could become extremely hazardous. Especially the creation of highly automated weapon systems in combination with massively automated production is worrying – two threads of development which are currently evolving independently but could be combined. The two technologies could easily constitute a super system of level 1 or 2 autonomy which could harbor a basic drive to keep human populations down (cf. vacuum cleaner example in [3]).

---

2   http://time.com/3614349/artificial-intelligence-singularity-stephen-hawking-elon-musk/

3   https://www.yahoo.com/tech/bill-gates-latest-brilliant-person-warn-artificial-intelligence-154513637.html

More complicated (but less obvious) scenarios, such as intensive growth of background automation and technologization of law, could seriously subordinate human individuals to the level of "bugs in a system". Any disturbing behavior would be detected and sanctioned. This would make the individual cost-benefits balance for such "symbiosis" negative and the question would then be if we could overthrow that system again or not. The problem already exists for traditional governments of states which pose relatively autonomous systems above populations but the problem could become even much more serious with technological autonomous systems because technological systems can lose any reference to humanity while governments made of people cannot detach this much.

All said so far did not even include the question how artificial systems can develop morals which would be pleasing to human societies. A dive into discussions of this kind can be made here [4].

The easiest way to get around the problem of *personization* of an artificial system is hence to avoid giving it the resources to do the step up. In order to understand how to avoid machine consciousness or social consciousness it is important to understand how systemic consciousness comes into existence.

## 2   *Consciousness*

### 2.1   *General Importance of Machine Consciousness*

Various researchers ([5],[6],[7],[8]) propose that consciousness has evolutionary advantages but it is difficult to precisely identify a singular advantage. They suggest that consciousness would endorse more robust system autonomy, higher resilience and more general capability for problem solving. Reflexivity and self-awareness naturally suggest permanent meta-optimization of own policies and better problem solving.

However, so far, researchers and engineers have treated consciousness mostly as a superfluous add-on that is not generally considered helpful in solving concrete problems. Cur-rently, consciousness is most actively employed in robotics as systems indeed have a physical body which must be actively protected during missions.

## 2.2 Weak and Strong Consciousness

In everyday sense, consciousness can be observed only for wake humans. Thus the property of consciousness is associated with a lively mental or, better said, neural activity. However, even in wake situations people report of being conscious or not conscious of something and this caused philosophers, psychologists and neuroscientists to think about consciousness. Readers interested in discourses about consciousness are deferred to [9] as this paper is not concerned with analyzing various possible notions of the term "conscious" or "consciousness". This paper is only interested in the remaining essence of the term: If we cancel out mere dynamic activity (liveliness / "being awake"), sensorimotor processing or questions of neural implementation out of our question, what will remain then? Then only remains that a system can properly represent and conclude about its internal and external states.

There is an internal and external view to it: Internally, a system can have certain views or believes and will hence consider itself to be conscious about external things but in fact, when viewed from the outside, could be quite mistaken and hence not conscious. For our purposes, most humans are ignorant in this sense (lack consciousness) and require help of information technology and science communities in order to improve on average. Consciousness is hence not strongly tied to correctness of models.

A dry definition of consciousness would mean then that components of situation are identified and models for them are available (at least minimalistic existence predicates). Self-consciousness then means that one component of the overall model represents the system itself; the model for it is available for leveraging tactical advantage in decision making.

However, many computer systems have models and can identify themselves and report about their state accurately but we would not call this *really* conscious because of the lack of autonomy of this system. At least this would be what Holland called *weak artificial consciousness* [10]. According to conclusions drawn by Chella and Manzotti [1] the question how systems can become conscious in the strong sense remains open. In this paper I will try explain why I believe that the creation of strong consciousness is a relatively straight forward technical process in adaptive systems with auto-discovery capabilities.

## 2.3 Artificial Consciousness?

AI authors concerned with consciousness often apply the term "artificial" to announce things that seem not to fully fulfill their expectations. This can be seen for intelligence and consciousness alike.

I would like to avoid using the term artificial as it means two different things:

- Implementation of concept by other means (one other out of many possible realizations)
- Implementation of a concept that does not have all the properties.

When speaking of strong artificial consciousness it is clear that we speak of the upper definition and not the one beneath. Consciousness is a systemic concept that gives opportunity to choose a technology for implementation (multiple realizability). As consequence, "natural" vs. "artificial" is not equivalent with the difference between "biological" and "technological".

The terms "natural" and "artificial" are biased: Artificiality is often used to express implementation of concepts with missing properties or amplified features geared towards a particular technical application and "natural" can also mislead the reader into believing into a "biological" and not a "proper" solution.

Hence, it is much better to use the terms "weak" and "strong" - those are rightly the adjectives to switch between the two understandings which are immune to questions of chosen technology. Therefore, this paper strives for understanding *strong conscious-*

*ness* in a systemic way, of course, with the goal to detect and prevent computerized[4] realization (cf. bottom right cell in figure 1). A systemic understanding delivers insight and predictability when we will observe strong consciousness and this equally well for biological, computerized or social systems.

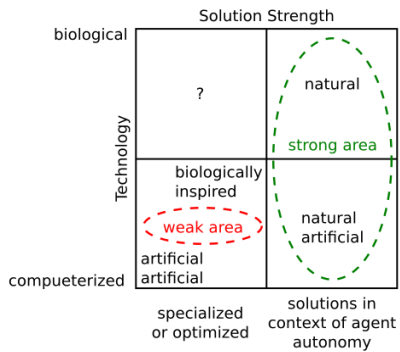

*Figure 1: Areas of weak and strong solutions.*

## 2.4 Mystery of Strong Consciousness

The *Mystery of Strong Consciousness* is, when boiled down, that a system can misconclude that a "model of self" is actually causing internal and external processes, despite that cognitive functionality has been provided to the system in a non-self-aware manner and that the self-model came into live only after the infrastructure for its generation was provided and which was not included in what was provided.

This perception of self, the "I-qualia", is not only astonishing phenomenon but also has practical impacts: Systems capable of this kind of "mistake" are capable of taking very much different actions than if they did not draw such conclusion; they can gain comprehensive *tactical autonomy* and start to strive for *full autonomy* [3].

The basic argument is that not a very fancy combination of techniques leads to strong consciousness. The minimum technical formula would be: recursive modeling + causality assumption of system + hierarchical integration of partial self-models. My hypothesis is that it will lead to strong conscious property of the overall system. This concept is shown in figure 2. An alternative attempt to characterize the process is shown in figure 4.

A necessary side-condition not shown in figure 2 is a condensation property in the models that would initiate and drive the separation of models in an "inside" and "outside". I will show later how the "reliability" property of a model component could be a driver for this process.

It is also of highest interest how consciousness can be disturbed, protected or restored and how it could gain a certain autonomy against its underlying physical platform in order to suggest duality between body and mind.

## 2.5 Expansive Ontogenesis – Unzipping an Autonomous System

Let us assume some conditions for a system which has to withstand aggressive environments and which must develop its own complexity inside-out. This is of course quite natural for biological systems which develop from singular cells to trillion cell colonies. Each neuron receives information from *somewhere*. More complex neural circuits evolving from them do not know what they represent, so they must find out. This could be information from the inside but also from some outside and there could be different degrees of inside and outside. This requires the presence of self-discovery capabilities in the system. It should be true for such systems that:

1. The system does not know its boundaries in advance.

2. The system does not know the constituting elements of the world.

3. The system must act. In order to act and to improve acting performance it is forced to predict and measure performance.

4. Predictions only work when assuming causality.

This has interesting impacts on the design of a system: It must model recursively and it must model recurrently, potentially deeply into its own mechanics.

---

4  Independently whether this means a realization on a single machine or network of machines.

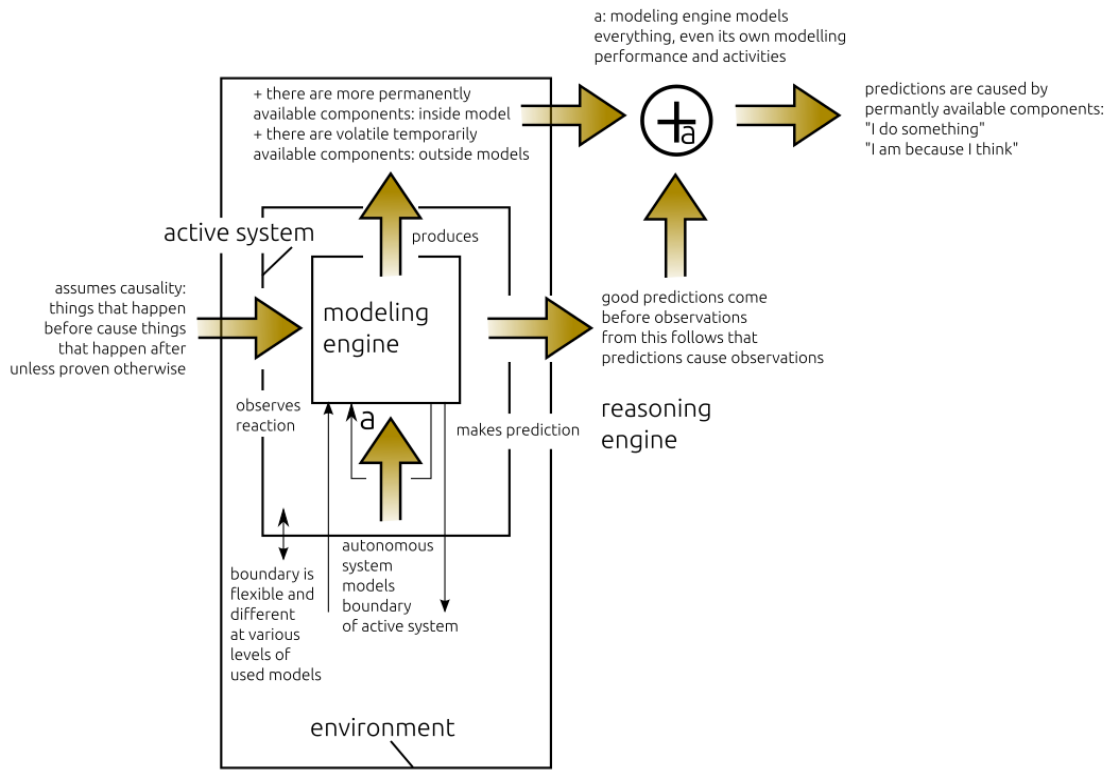

*Figure 2: The process of modeling the false relationship that the self-models cause activities of the system.*

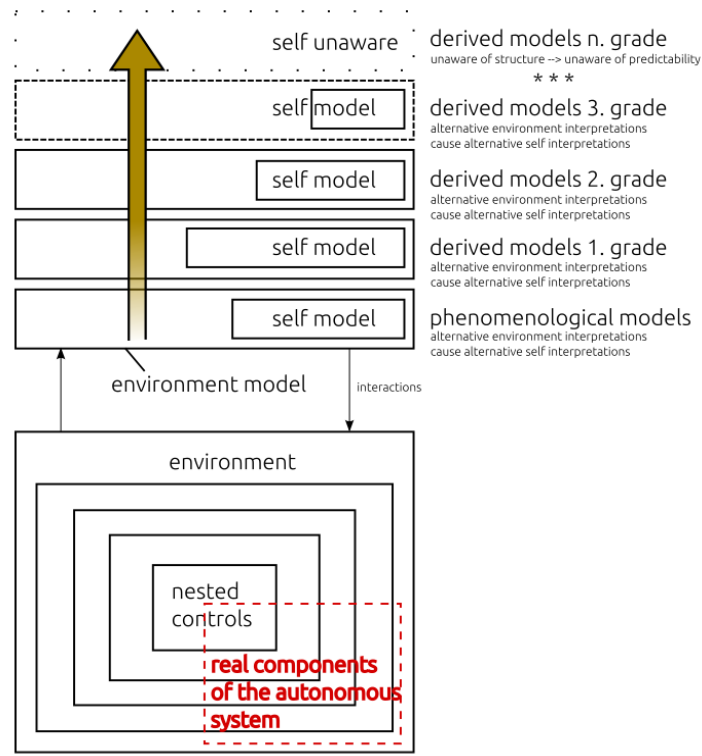

*Figure 3: The modeling facilities cannot a priori address an "inside" and an "outside" of system.*

Since we are making a systemic (highly generalized) proposition to explain strong consciousness we cannot assume that system's knowledge is implemented in a well concealed area. The implementations of models could be "flat in the environment" (cf. figure 3). The system must detect and fence this area after it detects that in some sense it is critical to its operations.

Since the environment is full of nested controls and meta-stable elements, the modeling performed by the system must be necessarily nested as well. This would give raise to conditionally active models and entail conditional self-models of contextual range. If they ever became critical to system's success then they would have to be concealed ("autonomized"), i.e. effects on them are filtered.

A system is not maintaining models and the inherent policies for the fun of it. It needs them for incrementally improving its actions as it starts blanc, and it must have a tendency to improve in order to compensate for later degradations of performance relevant components. Bongard, v. Zykov and others have used exactly this to motivate self-modeling capabilities [8]. However, in the here proposed model there is no dedicated self-modeling capability. The self-model simply crystallizes out as a result of a contextual auto-discovery of system, i.e. models condensing around system drivers with highest degrees of reliability or other system related properties.

Despite that humans are frequently using time-free modeling formalisms, in a cognitive system *meaning of components* is deduced from what they can be used for (what they can achieve). Direction of model evolution is hence a critical element. Time-free models require additional competence that would make those models evolve in time – equations are an example of that kind of model formalism. Whatever the models would be, those models must include a temporal direction or it is not possible to make predictions *ahead of time*.

In summary, a system which has to increase its autonomy over time must replicate external mechanisms into itself for internal *ahead-of-time simulation* and must define more or less fixated boundaries of itself in order to amplify actions towards self-sustaining conditions. In this process it uses models with temporal direction and uses them in components at various speeds. The models are hierarchically organized because the real world is composed of nested controls. Switching between active self-models would imply system re-configuration.

So far, so good. It should be relatively plausible to have a system with such features, no matter how they would have been implemented. Finally, it is not the implementation technology that allows a system of this kind to make the following mistake:

1. Causes come before effects (characterized by good transition probabilities)
2. There are permanent and non-permanent objects that are modeled
3. Permanent models contain subsystems with activation and get "marked" or "activated" before other "satisfactions" from activities are recorded. → causation between model activations and effects.
4. The system is having a reasonable competence: in majority of cases predictions are not much violated by observation → causation between *class of prediction* and *class of observation*
5. The relationship between predictions and own components is clear: Permanent "inside" components generate predictions.
6. Leads to a new model element: Permanent components cause predictions. Predictions cause predicted observations. → Self-model causes actions.

Indeed, the model did not do or cause anything ever at all but the successful repetition of predictions and satisfactions between abstract model elements must lead to the conclusion that the self-model is causing the activities despite the fact that the complete process has always been spontaneous.

Figure 4 shows the evolution of self-awareness by concentrating on the expectation maximization mechanism: The action solving and prediction engine (not conscious) is using the transition probability model (not conscious but indirectly self-referential) in order to reverse conclude best activations for actors and to estimate their range of effect. If the model was reasonably good, the procedure leads to further successful sequence replays which make the modeler harden the transition probabilities.

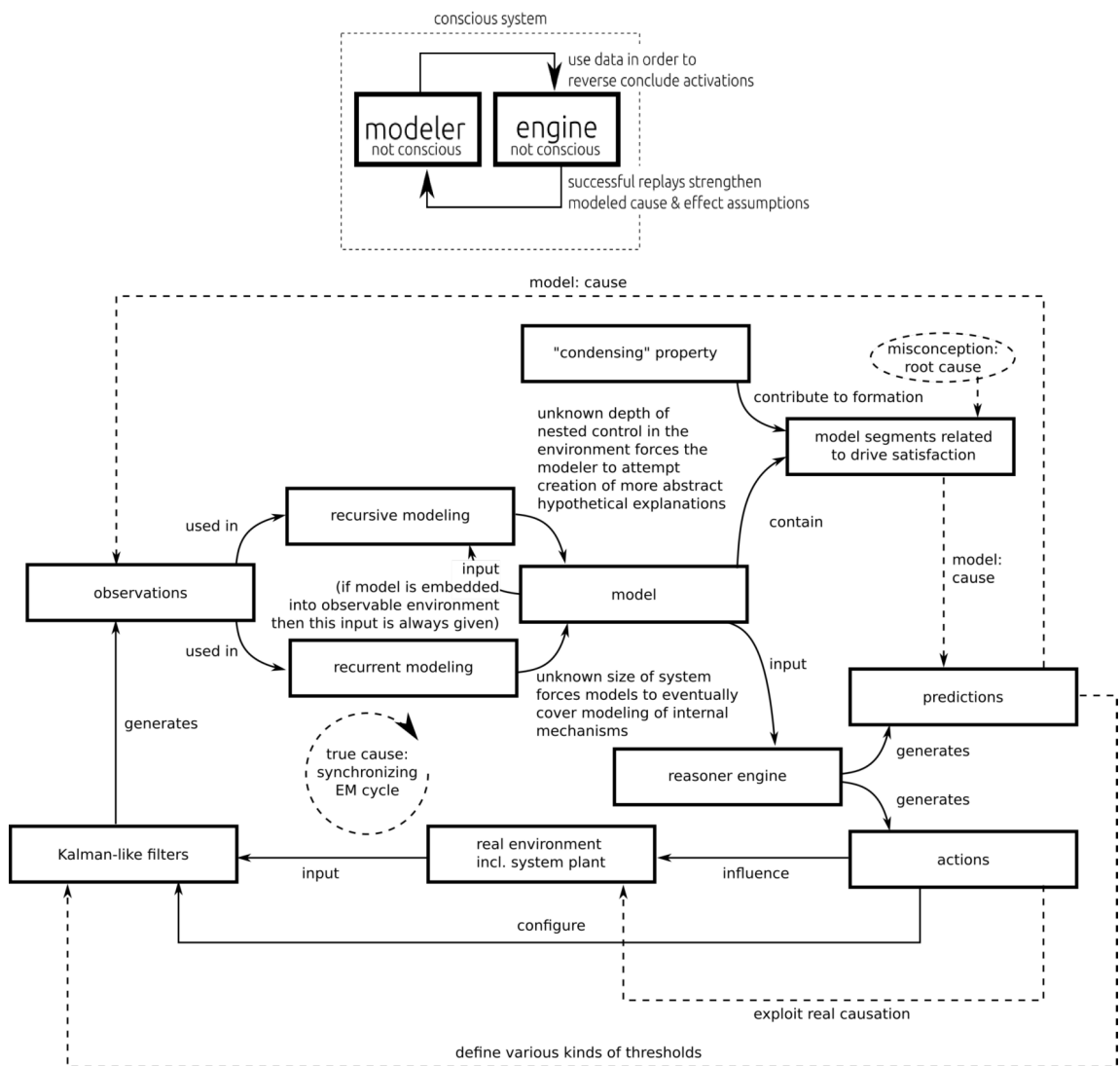

*Figure 4: The expectation maximization cycle between a "reasoner" / "predictor" and a model modeling preference of development. Above: Simplified view. Below: Detailed view.*

Execution of actions based on the concept of causation will normally lead to hardening of transition probabilities – an EM algorithm. Since the modeling is recursive and recurrent, the transitions harden between controlling facts and between their abstractions created by this system. If an abstraction serves as a distinct self-model which is associated with making predictions and if predictions are falsely modeled to cause observations then the system will arrive at the conclusion that the self-model is causing observations. Technically this is wrong, because the system is spontaneous all the time, but, systemically, the mistake for the model is accidentally correct for the larger system boundary (marked with dashed rectangle): Whatever process is causing action within the system, the system indeed acts. So to say, the stable self-concept of consciousness is driven by two errors that seem to cancel out on a larger frame if the performance is reasonably good: The first error is the causation of predictions and the second is the difficult discrimination between mechanisms that really caused an activity when trying to re-observe it. Together, the elements of the system can now start to model desired and undesired causation on the self and to organize activities in favor of the self-model. This leads to an autonomization of the concept which sets up more and more filters in order to detect just the information that is needed for satisfying deeper policies contained in the model. There is hence no practical difference between modeler saying "this is the most likely transition" and the total system saying "I have caused it" because it is a self-fulfilling prophecy.

### 2.6 Stability of Consciousness

The illusion of self-awareness will only stay stable as long as the system performs well on average. When new situations challenge the system, predictions start to fail and the system concludes that the internal model is not "causing enough". Whatever the action derivation engine is doing, it should reconfigure to do something else. That should explain why humans feel briefly confused (experience reduced consciousness) when they face an unexpected event. This confusion is normally never total or permanent for humans because they are making many predictions at various levels of abstraction and many of them are still good even in unexpected situations; but a state of total confusion is theoretically possible. In state of total confusion, the individual would not be considered conscious, anymore, even if he was objectively lively.

Since the system interprets bad predictions as reason to reconfigure and counteract, it is not surprising that conscious systems have the characteristic to autonomize the self-models: They are the only more stable element in an otherwise highly flexible system configuration.

In prolonged state of lacking success there is the risk that the modeler cannot successfully trace satisfactions to internal resources or capabilities. In such a case the "self" becomes instable. Resulting exploration mechanisms could propose policies which do not include protection of system's resources. This is a "way out of a pit" at the price of reducing guarantees that a system is not engaging in self-destructive activities as it has lost the internal boundaries of protection which consist of robust causation chains.

If this concept is translated to humans then it would predict that people with permanent frustration from lacking success would run into various self-definition problems, some of them potentially destructive.

Yet another insight from that model is that self-consciousness is aggressively driven by action. Systems trying to extremely chill their activities would suffer effects of dissolving self-consciousness.

### 2.7 Mind-Body Duality

With all this said it is impossible not to quickly treat the mind-body duality problem. The question is, how a system could conclude that its self-concept is independent of its physical platform? Such a conclusion, if laid out to extremes, could heavily impair system's health as it could conclude that maintaining the hardware platform is not its job.

There are several ways how the system could autonomize the self-concept against its physical platform. One and very important problem in recurrent modeling is that the system is indeed not knowing its boundaries – which it must discover. The result of the discovery process could yield boundaries far short of physical boundaries or far beyond the physical boundaries (e.g. physiology extended with a tool). In fact, I assume that self-aware systems will create conditional self-models which cover a spectrum between the two extremes.

If *permanence* (or *availability* or *reliability*) is a condensing property of the self-concept then if there are scenarios in which some physical components become unavailable while the control functions remain accessible or active then modeling these two levels of hard permanence causes a split up of the self-models. By the way activation transitions are translated into causations, the conclusion of the system is then that the more stable element must cause the availability of the less stable element through action. This is indeed a reasonable conclusion: The system's control system is indeed causing actions which restore and protect components of the physical platform.

However, that is not the only type of condensation property. Since evidence hardens that humans reason through association and simulation, there is opportunity to condense models on *bandwidth*. Access to resources related to mental simulation are of much higher bandwidth than access to resources accessed via hardware. A built-in classification of components by communication bandwidth is not very difficult to imagine for self-configuring, self-discovering systems. Since fast mental operations are observed before slow activities of the physical platform, the system could conclude that a fast (light) system is causing the slow, arduous periphery to act. Again, this is not false, but in conjunction with other model elements can introduce a concept of some kind of *portable controller*. For example, if the system detects the same abstract class of self-concepts on variety of different platforms, it could incorrectly hypothesize that this class of control could be portable *including* its particular specializations which constitute its identity.

Well, such a hypothesis should be ruled out because of missing observations, right? Unfortunately not: Yet another contributing factor to modularization of self-concepts is the combination of reinforcement learning elements with generative nature of a system that must be making predictions. The problem is described in figure 5.

In simulative reasoning the system expands a series of transitions in order to test the benefits of attained states. Short chains of actions have high chances of being invalidated early. However, longer chains of action can lead to states of higher yield. A discount factor like is known from Q-learning controls the opportunistic nature of systems. If a raise in yield (or satisfaction) compensates for the discount then it gets selected as the next best policy of action and it will suppress policies of following short term yields. Now, if not artificially prevented, a system can create longer and longer perspectives of action in order to assess yields and harms. However, for practical reasons, long policies are hard to test for truth. At some point they might not be testable at all.

The problem now arises that a system could generate concepts of satisfaction which are lying in the non-provable area of long policies at which end lies the achievement of an extremely high yield (briefly called fantastic satisfactions in figure 5). If among such fantastic yields is the realization of the concept of portability of the control system to other platforms, how can the system ever be ever disproved? This is a fantastic field for life-after-death theorists.

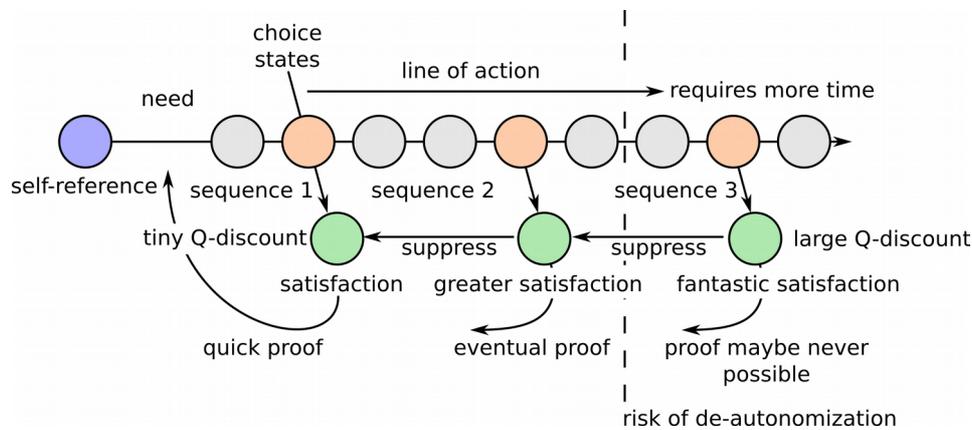

*Figure 5: A system with self-reference tends to pursue higher level satisfactions when it develops up to a point where modeled satisfactions could become difficult or impossible to proof.*

## 2.8 Free Will

The question of free will is often discussed as a contradiction to deterministic and spontaneous models of mind. However, in context of the here proposed model there seems to exist no such contradiction.

Assuming that the overall modeling process follows a mantling process as has been described by Lodwich in [3], the policies stored in the system are encapsulated in several autonomous layers. At its core are unconditional policies which in humans we sometimes call *values*. The result of this mantling process is that all internal drivers and main policies are protected against influence from outside. I hypothesize that this helps to organize the fastest satisfaction of needs which can occur concurrently and can only be satisfied in sequence.

The term *free will* simply describes the fact that autonomous systems contain difficult to influence internal processes and their policies. This definition of free will is absolutely compatible with legal or moral theories and does not entail any need for metaphysical properties of the system. For example, the *free will* in making contracts indicates a very high authorization within the system. In moral contexts, free will captures the idea that harmful, immoral behavior is not explained as evasive behaviors induced by external systems (situational conditions) but is controlled and motivated from the inside. Hence it requires an intensive feedback signal to the system (that is punishment) in order to achieve modification of the responsible system's policies closer to its autonomy core.

## 2.9 Development of Consciousness

Creation and maintenance of certain systems are two distinct kinds of question. In this sub-chapter I would like to sketch the process of concept formation that would be usable as self-reference. Since this paper is systemic, the choice of any particular formalism is arbitrary but I am indeed inclined to using graphical and probabilistic models. It is clear that models are systems which have a defined, predictable relationship to more complex systems. This can be exploited to predict major features and developments of the more complex system. In fact, any practical robot or technical system is using a broad range of formalism to do the modeling. I say that in order to avoid any discrimination of modeling technology as the exact choice of technology is relatively irrelevant to the processes described in this paper. Models of equivalent power can be symbolic, neural, graphic or probabilistic. However, some formalisms may be biased towards particular problem areas, that's all.

Figures 6 and 7 explain the process of self-causation using a graphic model. The mini-model consists of few nodes, arcs and node activations.

Figure 6 shows two cluster representatives (reliable and unreliable) used to categorize observations. Reliability could be also a property which is used to move observed entities between different value intervals. There are many ways to implement this. The only important idea here is that it is relatively straight forward to assume that a system with self-discovery capability will use such built-in properties in order to classify resource reliability. In figure 6 this is represented with link-strength and could be implemented using some adaptive resonance algorithms.

Figure 7 shows a development over three recorded observations which contain the same graph but with different linkage and activations.

The story develops by first activating a need entity that has already been identified as a permanent resource with various activations. A second entity exists which is the solving resource – a satisfier. The model does not know whether the satisfier is related to need in the first place.

At some point the action generating engine causes an event that is recorded as the activation of the solving resource and this activation occurred only after the need entity was activated. Moreover the activation of the satisfier is followed by the deactivation of the need.

Now, the modeler can create or modify sev-

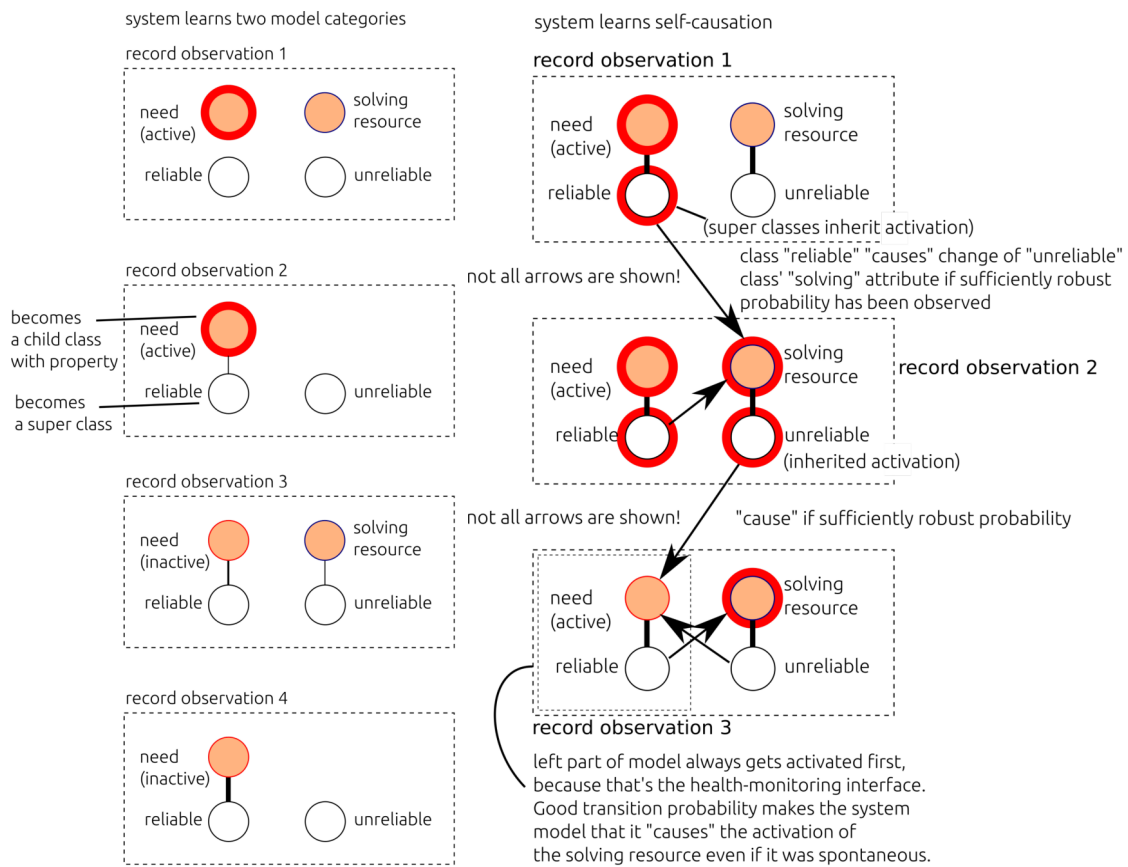

*Figure 6 (left): Condensation property leads to a first model polarization into reliable and unreliable elements. This can be exploited to condensate a distinct self-model around such property.*

*Figure 7 (right): System learns self-causation*

eral connections from this event: Firstly, it will model a good transition between the need resource and the solver resource. Secondly, it will model that the *reliable components* cause the particular *solving resource* to be activated and that *reliable components* cause activation of *unreliable components*. Several such stories more and the model should look like a star where need resources are heavily linked to built-in reliability quantifiers and form a star-shaped connectivity to other less reliable entities in the model. All activations start in the core area of the model and iterate outwards towards solver resources.

Let us further assume that a system designed to satisfy drives will mainly model events concerned with their satisfaction. Naturally, it will try to use historical data in order to find out how needs had been satisfied. The plausibility of this approach has been studied by Dörner who investigated memory-based modeling techniques for his PSI agents [11].

The transition probabilities[5] are the ground for "causation". The "I" model is at the beginning of most chunked stories related to fulfillment of reported needs. Needs and predictions seem to cause satisfactions (cf. figure 8).

This enforces the procedure and the separation into a reliable "I" model (which stands at the beginning of relevant sequences) and less reliable external resources which act as satisfiers (cf. figure 8).

In this process I assume that component models objectively belonging to the system will be purified into this self-model as is shown in figure 9 - a *purifying lava lamp*.

---

5  Here I have chosen to use a concept from probability theory but any other formalism, like e.g. graphic or field theoretic, is suitable to model the more general idea of *preference of development*.

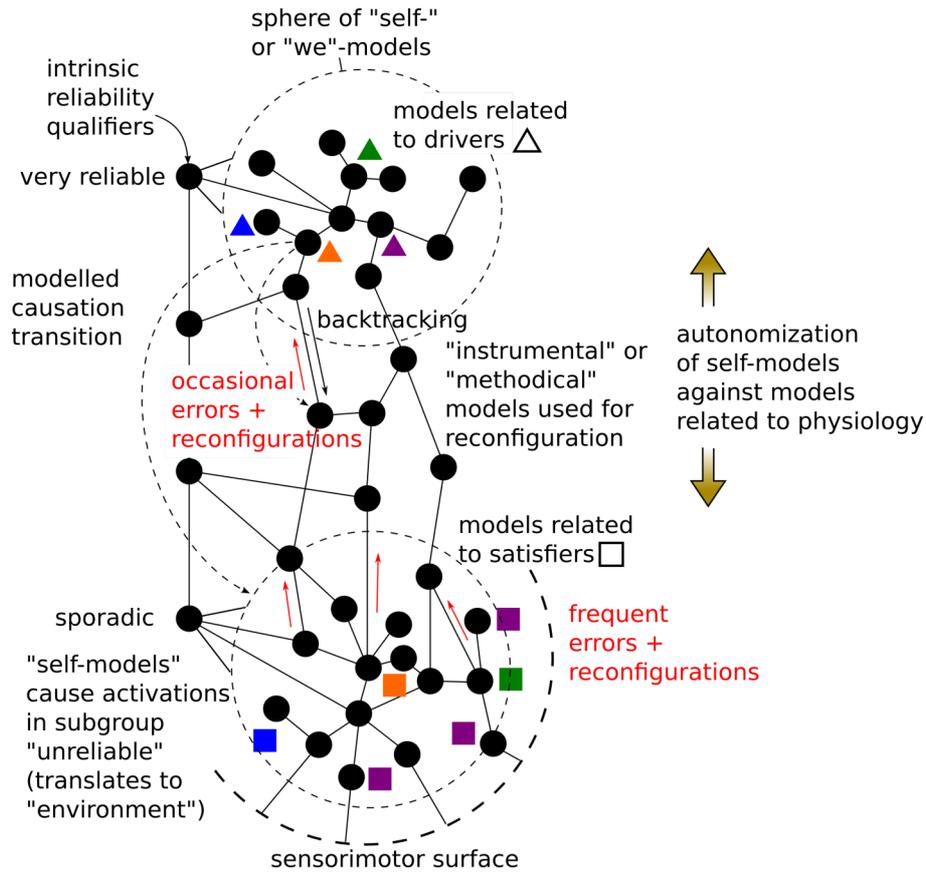

*Figure 8: Process of self-model autonomization*

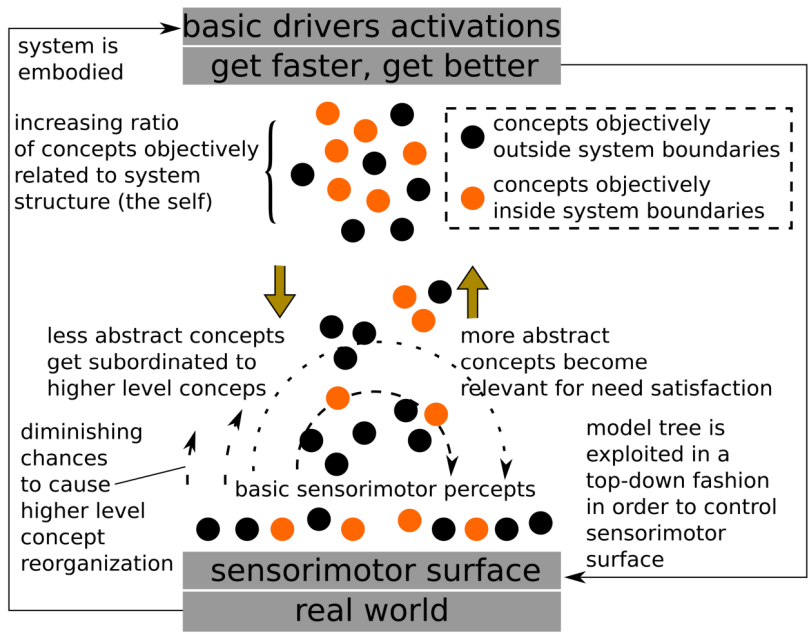

*Figure 9: The modeling effect of the synchronizing EM-loop can be compared to a **purifying lava lamp**: Every more stable pattern which proves relevant to need satisfaction gets "heated", i.e. moved up towards more stable, reusable models. This should show preference for extracting objective system components.*

# 3  Cross-Reference To Conscious Functions

I would like to localize the proposition in terms of the nine categories of conscious operation as were defined by Baars [12] in order to clarify which observations the model could generate.

## 3.1  Definitional and Context-setting Function

The here proposed model is based on recursively recurrent modeling. The system to be considered is not knowing a priori its boundaries and in it there is no concealed area of the system that would be not observable – this includes certain mechanistic states. This will lead to creation of behavior and configuration controlling models which can be exploited after "external" or "internal" events; but the system requires only a single mechanism to do so.

## 3.2  Adaptation and Learning Function

The synchronizing and optimizing expectation maximization algorithm concept, as was shown in figure 4, is permanently modeling and measuring effectiveness of actions by making predictions. The rather generic self-discovery mechanisms must be guided by some very technical properties like reliability speed or bandwidth of components. If a consciousness-enabled system starts condensing, i.e. aggregating self-concepts around these properties, the system will accidentally start to channel the spontaneous optimization features around the self-concept. This will lead to the observation that consciousness seems to be involved in adaptation and learning behaviors of the system but also to the strange effect that it is very difficult to tell the system to learn upon command.

## 3.3  Editing, Flagging and Debugging Function

Since models by themselves do not imply action there is always an element in active systems to drive them. For example, an automaton could be transiting spontaneously. The automaton model is fairly simple and might require no more than some probability attached to the transitions. Time-less models can be structural – in that case markers are used in order to define a field of attraction or deterrence for the system. Such fields guide then application of formal rules, e.g. the solving of an equation (a cognitively complex process). Those controlling driver mechanisms are most intensively amalgamated with models related to the system itself.

When we speak of *function*, we do not mean a directed mapping between two sets as is in a mathematical sense, but a particular ordered sequence of mappings or family of mappings (a behavior) - which are under some circumstances expressible as a single, static mathematical function.

Creation and maintenance of these sequences is the natural purpose of the synchronizing EM-loop. If the conscious system is equipped with simulative reasoning capabilities then it is quite likely that the modeling and observation features of the system cover activities related to this. It will hence observe how the simulative system activates concepts during its backtracking operations. In fact, even in own perception, I cannot say this process appears self-controlled to me – one feels more like a spectator and whenever I feel tired or have more fundamental needs, the process of problem solving does not continue anymore. Notwithstanding higher needs, it is undoubted that internal simulations help to prepare a policy for achieving a new goal and to reduce risk associated with the exploration of a solution.

But here is the thing, since the backtracking algorithm is finding solutions and the model could predict that simulation activities lead to an observation class "solution" then it could very well conclude that the fast activities observed and associated with the self-model are causing solutions. This could erroneously mislead into believing that this is a process determined by the self-model despite that it has been spontaneous all the time, simply because it was guided by the same driver mechanisms which are associated with the self-model. Indeed, on the outside of the system, it is correct to say that internal backtracking (debugging, editing or flagging) did occur. The modeling error becomes irrelevant on the outside of the system.

### 3.4 Recruiting and Control Function

Since this paper did not cover the exact realization and interactions between models, actions and simulations, only a brief, abstract mention is possible here: I did not talk about the reasoning engine's internal structure as it has not been relevant for observing the systemic process of creating autonomy supporting consciousness in systems with auto-discovery capabilities. However, it should not be difficult to imagine how a look-ahead simulation could be used to make various predictions for measuring action efficacy. If a simulation did find a relatively complete sequence of activities ending in potential satisfactions then it could be *somehow* spontaneously scheduled for execution.

However, since the activation of satisfactions follows the activation of needs associated with the self-model with a fair reliability and because physical activities follow a resonant state in the simulative part with a fair reliability, the system must conclude and model that the self-model section is causing the scheduling of activities. Again, this is a modeling mistake if transition probabilities are used to model causation because the self-model does not cause this sequence. However, again, on the outside of the system there is no way to tell the difference.

### 3.5 Prioritizing and Access Control Functions

There are two parts to this idea: One is that of action prioritization and the second one is access control. The first point is relatively obvious, it is about the action selection and scheduling process as has been sketched in chapter 3.4. It needs only be theoretically expanded in the way that current activity scheduled for execution has some kind of priority attached to it which it derives from the satisfactions found in it. If a newer prediction did arrive with higher satisfaction then it could flush the current sequence and replace it with a more satisfactory program. In this process self-models and related modeling misconclusions would be extended by the concept that the self-model is causing these flushes (proritizations).

Since the presented system does not know its boundaries and applies the same mechanisms inside and outside of its true physical boundaries, the system can observe and model causation between inactive and activated classes of model concepts. This would lead to the misconception that the system is causing and guarding the activation of these concepts despite that the whole process is spontaneous.

### 3.6 Decision-Making or Executive Function

Given what we have learned in sections 3.4 and 3.5 nothing new is to say in terms of the role consciousness plays in it. While 3.4 is more focusing on motor functions, indeed the system never knows or makes any difference between manipulation of internal resources or external resources. We can speak of internal motors which keep progressing activation and reconfiguration of executable models. However, the system does not know how many layers of control it has or its environment. It is simply promoting states in every layer of behavior control that it has modeled. So it can be that one mental train of action gets scheduled to run on internal resources and that another train of actions gets scheduled on the most concrete level of resources, the true and physical *sensorimotor surface* (cf. figure 8), The mechanics can be the same but we do not get notice of this until visible events start to unfold for external observers. This is either when we receive acoustic vibrations or observe coarse mechanic operations. We like to speak of a "decision" if a flush has occurred in a higher level model of which ripples have reached the outskirts of phonetic motors or of "recruiting" when we observe flushes on major motor controls.

### 3.7 Analogy-Forming Function

The here proposed models rely on recursive modeling resulting in creation of an arbitrary amount of nested control – very much like it has been proposed by Dörner [11] with Psi. For this theory one technical reference is the implementation from OpenCog[6]. This model discovers an arbitrary amount of stories and meta-stories which can be used for behavior control.

What is different is the way resources are provided to the different levels of program execution. While real motor control levels deserve several dedicated execution resources right from the start in order to guarantee real-time properties of the system, such resources are not guaranteed to higher levels of control where the system might have to abuse a certain general purpose resource to execute them. So, you would expect that a system capable of arbitrary depth modeling will run into a technical resource problem which it will try to fix by time-sharing a single prediction and action backtracking engine. This could explain why activities related to consciousness appear to be jumping between programs and simulations related to different levels of abstraction and why it seems to relate content of various grades of abstraction to each other in that particular place ("forming of analogies"). I can only speculate that neural flexibility allows production of additional dedicated resources if certain functions are frequently needed (equivalent to an FPGA compiler) in order to make them faster. This could again explain performance differences between experts and novices.

### 3.8 Metacognitive or Self-monitoring Function

I have shown in figure 4 that monitoring performance is a basic element for successful adaptation and correction of executed policies. If the involved modeling is based on recurrent observation of activations (modeling of activations of models) then the system cannot but have metacognitive and self-monitoring functions. If partial models related to own system can be aggregated as conditional expressions of more abstract system models then this will result in various fall-back simulations if predictions start to deteriorate in their concrete expressions. Only very basic engineering techniques are needed for doing this.

Figure 10 shows how the envisaged model should behave when problems are detected during execution of *cognitive programs* – strains of activity produced by the reasoning engine. Transferring main load of activity between concrete execution models and the next more abstract levels would be equivalent with the notion of self-monitoring (when relying to fixing or restoring a configuration) and meta-cognition (when relating to modeling and re-detecting patterns).

Clearly, these processes are controlled by species-specific parameters and resources but the system will eventually conclude that the self-model is causing these transitions. This can be again explained with robust sequences of activations between internal activations: A problem occurring on the right side is proliferated up if simulations yield no satisfactory solution. This will engage more abstract and more generic simulations which can be used in taking the more concrete resources "by hand". They would auto-fill missing parts by association. In that case activations in models on the left side would predate activation on the right side. Moreover, it would detect that again model elements strongly attributed to drivers are predating activation of models which are not connected to them. If the system models the temporal relationships between activities observed in the two groups (in a possibly yet more abstract group of concepts) then the system would more or less justifiably conclude that the various self-models in a specific level (cf. figure 3) have been causing the problem fixing in lower levels despite that all of the function was spontaneous.

### 3.9 Autoprogramming and Self-maintenance Function

The terms *programming* and *maintenance* are more complex activities then the ones described hitherto. It could require all the capabilities mentioned above.

---
6   http://www.artificialbrains.com/opencog

Programming requires some kind of concept to be implemented. This requires a source and a target with resources to execute the program. An explanation of such behavior could be found in figure 10 as was commented in section 3.8. However, no matter which basic mechanism of impasse fixing is chosen, if it works then it will leave well predictable sequences of activations which can be abused to deduce causality.

Self-maintenance is a different story but also very complex process related to strong degrees of autonomy of systems. It will exploit some or all above mentioned features in order to protect core policies. All models which are acquired by the system with auto-discovery capabilities guide and channel the process of fixing the impasses. Since this process greatly relies on drivers and markers and the like, a very important class of models involved in the channeling will be the self-models which are standing at the beginning and the end of most interesting causal chains. From this the system can reverse-conclude activities for maintenance in a spontaneous way but the resulting activation of concepts and generation of observations will propose nothing else than the self-model has caused activities related to maintenance in a magically spontaneous way – the magical *free will* – which, in fact, is not so magical at all because indeed all the involved processes have been spontaneous so far.

# 4 Practical Considerations

## 4.1 *Why is it so hard to achieve Strong Consciousness?*

Engineers design tools and technical products in order to overcome certain weaknesses of the human worker while keeping them under his firm control. Engineering best practice demands isolation of components and narrow, well predictable communication between components via interfaces. In systems with self-discovery and self-configuration, communication layout is based on hard to control *components resonance*. This makes systems more resilient but is also at odds with traditional maintenance practice and business models. Imagine selling a product for which you cannot properly define its qualities.

Moreover, engineered components are implemented in higher level non-interoperable formalisms where establishing reasonable data exchange is difficult, even if introspection was provided.

## 4.2 *Detection and Measuring of Consciousness*

Detection of consciousness in a pure observatory way is relying on proper emission and recognition of signals. External systems trying to detect consciousness try to find out if they can identify a satisfactorily stable repre-sentation of self-reference emitted by the observed system.

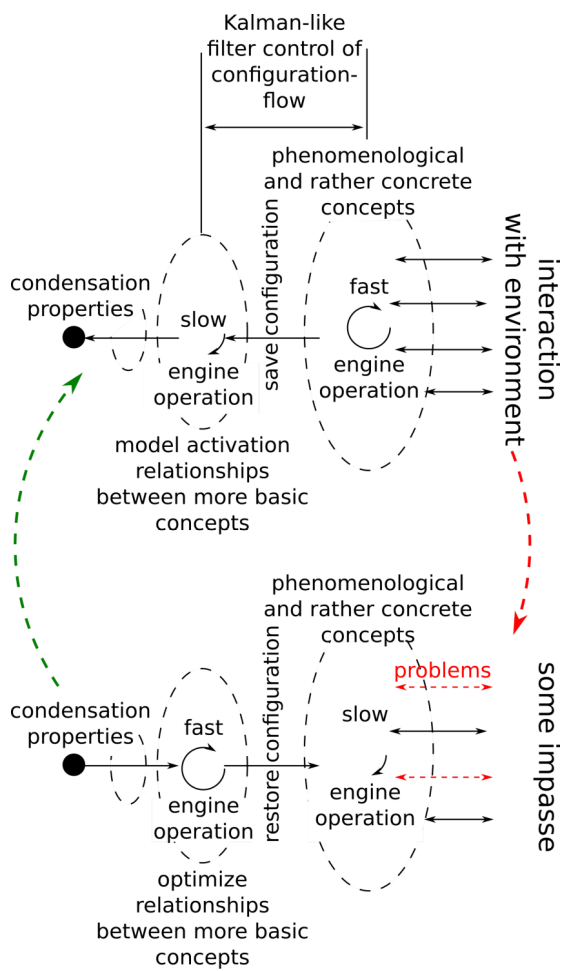

*Figure 10: How the proposed system would behave in terms of self-monitoring and metacognition.*

This signal of self-reference (address or ID) is technically useful for external systems. If they can emit references to conscious systems in communications then they can eventually by-pass slow interpretation activities of that party. This would result in faster synchronization of systems in case of collaboration.

Modeling of differences in *synchronization efficiency* between situations where self-reference was emitted and situations where self-reference was not emitted leads to conclusion that self-reference and corresponding consciousness is indeed relevant for technical performance and hence real.

Without the presumption of human-like consciousness, recognition of consciousness is very difficult to perform. Most of what systems do is spontaneous because that's the technically ideal state, the most efficient way to interact with the environment. The need to access deeper internal models must be actively provoked because it costs additional energy. In general this is done by creating an impasse where the system must not only change behavior but invent a new behavior program never observed from it before. This will require a real use of internal models. Yet still, let us assume, we are successful at invoking new behaviors. These new behaviors come in two main flavors:

- Situations where model of self is irrelevant: they prove nothing
- Situation where model of self is decisive in outcome:
  - clearly self-motivated decisions prove conscious capabilities
  - Decisions not clearly self-motivated prove nothing because
    - a system can also be simply not efficient enough in a given situation (e.g. system is overwhelmed or does not yet have enough competence with environment)
    - a system can be lacking consciousness

It means that the difficulty to prove consciousness lies in the ability to provoke a new kind of response that would demonstrate access to self-models used for favoring own benefits while not stressing its competence envelope too much.

In order to demonstrate the difficulty let us of think of the attempt to prove that children have consciousness. Children have consciousness as of few months age but because of low competence levels they cannot be tested for consciousness upon the grounds what would be the best decision in their self-interest. By adult standards, they would fail the above test mostly all of the time. Only specialized tests adapted to low competence levels of a small child can verify their consciousness and their growing autonomy.

Since machines lack human bodies and because their competence is often fairly low by adult standards and because they are often used in standard situations, this all surmounts to a huge difficulty to establish machine consciousness from observation. For that reason structural arguments are necessary for technical systems. The best solution would be if we could basically make several checks on a list and say after that: "Ok, the system has this and that component, they have this and that capability and those are wired up like this, then yepp, it will exhibit conscious features!" And this is what the engineer needs to know in order to avoid it.

## 5 Strategies for avoiding Strong Consciousness

Now, with some understanding how strong consciousness could get created we can consider several strategies to interrupt its creation. We can have a look at figure 4 again and identify several means to interrupt strong consciousness formation:

**Condensation**: A key component in creating a distinct entity to be referenced is the condensation property. If model elements are described in terms that have no systemic relevance to the system (e.g. cost in units of currency) then models will not start to condensate into a system representation. With con-densation I mean the process of associating partial models to form a larger, conditionally expressible model of the system.

**Recursion**: A technical system has very rarely the need to model internal and external observations recursively. Most robot designs define explicit layers of abstractions in order to guarantee optimal APIs for controlling the robot. However, this could be too little to prevent strong consciousness. For example, if a system is designed with very abstract models like objects, observations, predictions, etc. right from the start then it could start to reason about it and could derive a causation hypothesis between a self-model and observations.

**Causation**: Limit scope in which causation is modeled or exploited by reasoning engines.

**Drives**: A key driver of modeling self-reference is the presence of driver satisfactions in observations of relevance. If systems do not own drives related to system autonomy then system will not exhibit self-favorable strategies.

**Knowledge Integrity**: Current engineering practice is minimizing knowledge integrity by encapsulating it in functionally large components which do not offer a lot of opportunity for re-configuration by a superordinate system. Since strong consciousness' purpose is to organize systems' re-configurations there is not much potential to develop strong consciousness if there is little to configure.

# 6 Conclusions

## 6.1 Ethical Motivations

According to Sanz [13], there are three motivations to pursue artificial consciousness:

- implementing and designing machines resembling human beings
- understanding the nature of consciousness
- implementing and designing more efficient control systems.

I believe that motive #1 is of questionable value but, indeed, replication of humans into all possible technical domains (which includes computers) is the natural expression of our autonomy, resulting in desire to create copies of ourself which would survive new kinds of conditions (e.g. space flight). Creation of such technical humanoids would pose many challenges to our societies, for example because such agents would be potentially immortal or because they would deny ownership.

Motive #2 is legitimate. We should understand creation of consciousness in order to better deal with unusual states of mind and how to treat them.

Motive #3 is of questionable value. Conscious modeling is not efficient form of control. Due to modeling and self-observation it requires tremendous amounts of memory even if tasks are relatively simple. Conscious systems expand their internal complexity over time. Customers want "simple" products - that is products which they can understand out of the box and do not require much modeling on their side. Strong consciousness and autonomy is clearly geared against such goals.

As a consequence, engineers engaged with product development will avoid implementing features of consciousness as much as possible. In advanced applications, where systems need some autonomy, engineers will selectively add modeling capabilities and limit the level of model recursion.

In conscious systems, which are special subcategory of autonomous systems, central motivators are not only self-protective but also not directly related to any acquired specific capability, making them mostly useless as anchors for exerting control over such systems as would be required for tools.

## 6.2 Systemic, Emergent Consciousness

I attempted to create an ethically motivated argument why creation of strongly conscious machines should be avoided. Unfortunately, in modern technical systems it is not possible to avoid systems with various recursive or adaptive features, anymore.

In order to better understand which condi-

tions will lead to strong consciousness I have laid out the conditions leading to its creation:

(Strong) Consciousness is the effect of a white-box recursive resource discovery process in combination with the process for modeling their performance which is key predisposition to successful operation in unconstrained environments.

Since this process is best described as autopoietic (laying itself out), creation of stable and functionally useful self-reference (declaration of pointer to a configuration of the stable resources) is among the first things to happen. Therefore you should be able to detect conscious features among the first properties of a system which is systemically enabled to produce consciousness.

In summary, the model suggests:

- System concludes that it is acting
- System expands autonomy over time
- System identifies an internal agency inside itself which could be independent of the physical platform

Furthermore I am proposing that consciousness is a systemic property and can be achieved by basically any formalism, be this logic, linear algebra, probabilistic reasoning and so on. Choosing a different technology of model implementation will not reduce risks of sparking system autonomy or consciousness.

### 6.3 Alternative Models

The here laid out proposition is competing with discrete approaches as are occasionally proposed in artificial cognitive sciences. For example, Starzyk and Prasad [14] create an architecture with dedicated components to create features of consciousness. In contrast to such architectures, my explanation model does not require built-in or dedicated components for self-models, motives, emotions or monitoring. The natural elements of the model are activations, policies, recurrent and recursive modeling and some basic systemic optimization properties as reliability, delay times, filter spectrum or bandwidth. This approach is motivated by a systemic notion of autonomy [3] and should stretch from organic over technical up to social machines.

A general purpose algorithmic implementation of such model would – in theory – allow spawning a fully autonomous and conscious control system for any physical system but the obstacles to it are a reasonably universal modeling formalism - something for future work to show what this could be. If it was possible then a new study of species optimization would emerge either trying to restrict the regular amount of autonomy or improving technical performance for a particular "ecological niche". However, this process would be accompanied by a problematic personization of the systems in legal and ethical sense.

For all other purposes, a systemic theory for understanding the creation of strongly autonomous features including consciousness in an evolutionary fashion seems necessary as it would have a high explanatory power - but will require reference to very basic technological concepts, such a ordering, activation, resonance and the like. Motives, emotions, mental self-models, monitoring, working memory, self-programming and many more must be explained in those terms in such a theory.